\documentclass[letterpaper, 10 pt, conference]{ieeeconf}
\IEEEoverridecommandlockouts
\usepackage{times}

\usepackage[nolist,nohyperlinks]{acronym}
\usepackage{amsmath}
\usepackage{amssymb}
\usepackage[english]{babel}
\usepackage[T1]{fontenc}
\usepackage[utf8]{inputenc}
\usepackage{multicol}
\usepackage{hyperref}
\usepackage{siunitx}
\usepackage{svg}
\usepackage{tikz}
\usepackage{url}
\usepackage{caption,setspace}
\usepackage{array}
\usepackage{booktabs}
\usepackage[linesnumbered,ruled]{algorithm2e}

\usetikzlibrary{calc}
\usetikzlibrary{decorations.markings}
\usetikzlibrary{shapes,arrows,positioning}
\usetikzlibrary{circuits.ee.IEC}
\usetikzlibrary{calc}
\tikzstyle{block} = [draw, fill=blue!20, rectangle, 
    minimum height=3em, minimum width=6em]
\tikzstyle{sum} = [draw, fill=blue!20, circle, node distance=1cm]
\tikzstyle{input} = [coordinate]
\tikzstyle{output} = [coordinate]
\tikzstyle{pinstyle} = [pin edge={to-,thin,black}]

\newcommand{\norm}[1]{\left\lVert#1\right\rVert}

\DeclareMathOperator*{\argmin}{argmin}

\begin{document}

\title{A Fully-Integrated Sensing and Control System\\ for High-Accuracy Mobile Robotic Building Construction}

\author{Abel Gawel\authorrefmark{1}, Hermann Blum\authorrefmark{1}, Johannes Pankert\authorrefmark{2}, Koen Kr{\"a}mer\authorrefmark{2}, Luca Bartolomei\authorrefmark{3}, Selen Ercan\authorrefmark{4}\\ Farbod Farshidian\authorrefmark{2}, Margarita Chli\authorrefmark{3}, Fabio Gramazio\authorrefmark{4}, Roland Siegwart\authorrefmark{1}, Marco Hutter\authorrefmark{2} and Timothy Sandy\authorrefmark{2}\thanks{\authorrefmark{1}Autonomous Systems Lab, ETH Zurich}\thanks{\authorrefmark{2}Robotic Systems Lab, ETH Zurich}\thanks{\authorrefmark{3}Vision for Robotics Lab, ETH Zurich}\thanks{\authorrefmark{4}Gramazio-Kohler Research, ETH Zurich}\thanks{Luca Bartolomei, Margarita Chli contributed to the motion planning \tt\small \{lbartolomei, mchli\}@ethz.ch.}\thanks{Hermann Blum, Abel Gawel, Roland Siegwart contributed to the state estimation and high accuracy localization \tt\small \{blumh, gawela, rsiegwart\}@ethz.ch.}\thanks{Timothy Sandy contributed to the state estimator, and motion control \tt\small tsandy@ethz.ch.}\thanks{Koen Kr{\"a}mer, Johannes Pankert, Farbod Farshidian, Marco Hutter contributed to the Motion planning and control \tt\small \{kokraeme, pankertj, farshidian, mahutter\}@ethz.ch.}\thanks{Selen Ercan, Fabio Gramazio contributed to the building task interface \tt\small \{ercan, gramazio\}@arch.ethz.ch.} \thanks{This work was partially supported by the Swiss National Science Foundation (SNF), within the National Centre of Competence in Research on Digital Fabrication and by the HILTI group.}}

\maketitle
\begin{abstract}
We present a fully-integrated sensing and control system which enables mobile manipulator robots to execute building tasks with millimeter-scale accuracy on building construction sites. The approach leverages multi-modal sensing capabilities for state estimation, tight integration with digital building models, and integrated trajectory planning and whole-body motion control. A novel method for high-accuracy localization updates relative to the known building structure is proposed. The approach is implemented on a real platform and tested under realistic construction conditions. We show that the system can achieve sub-\SI{}{\centi\meter} end-effector positioning accuracy during fully autonomous operation using solely on-board sensing.
\end{abstract}

\IEEEpeerreviewmaketitle
\section{Introduction}
\label{sec:introduction}
The available workforce in the construction industry is decreasing, and safety regulations are rising, causing a strong demand for innovative solutions on construction sites, e.g., digitization of the construction process and utilization of robots. However, robotic fabrication is traditionally associated with automated assembly lines, where fixed positioning and constant conditions determine the role that the robot may undertake in the fabrication process. Unlike such stationary robotic facilities, construction sites are spatially complex and cluttered environments. A construction robot must localize in its environment reliably, while being able to work with conditions deviating from as-planned building information. While humans use established tools to achieve required \SI{}{\milli\meter} accuracy in construction tasks, mobile robots usually achieve only \SI{}{\centi\meter} relative accuracy using on-board sensing~\cite{ardiny2015autonomous}. This is not sufficient for many construction tasks such as drilling of holes, and installation of fixtures. 
\begin{figure}
    \centering
    \includegraphics[width = 0.9\columnwidth]{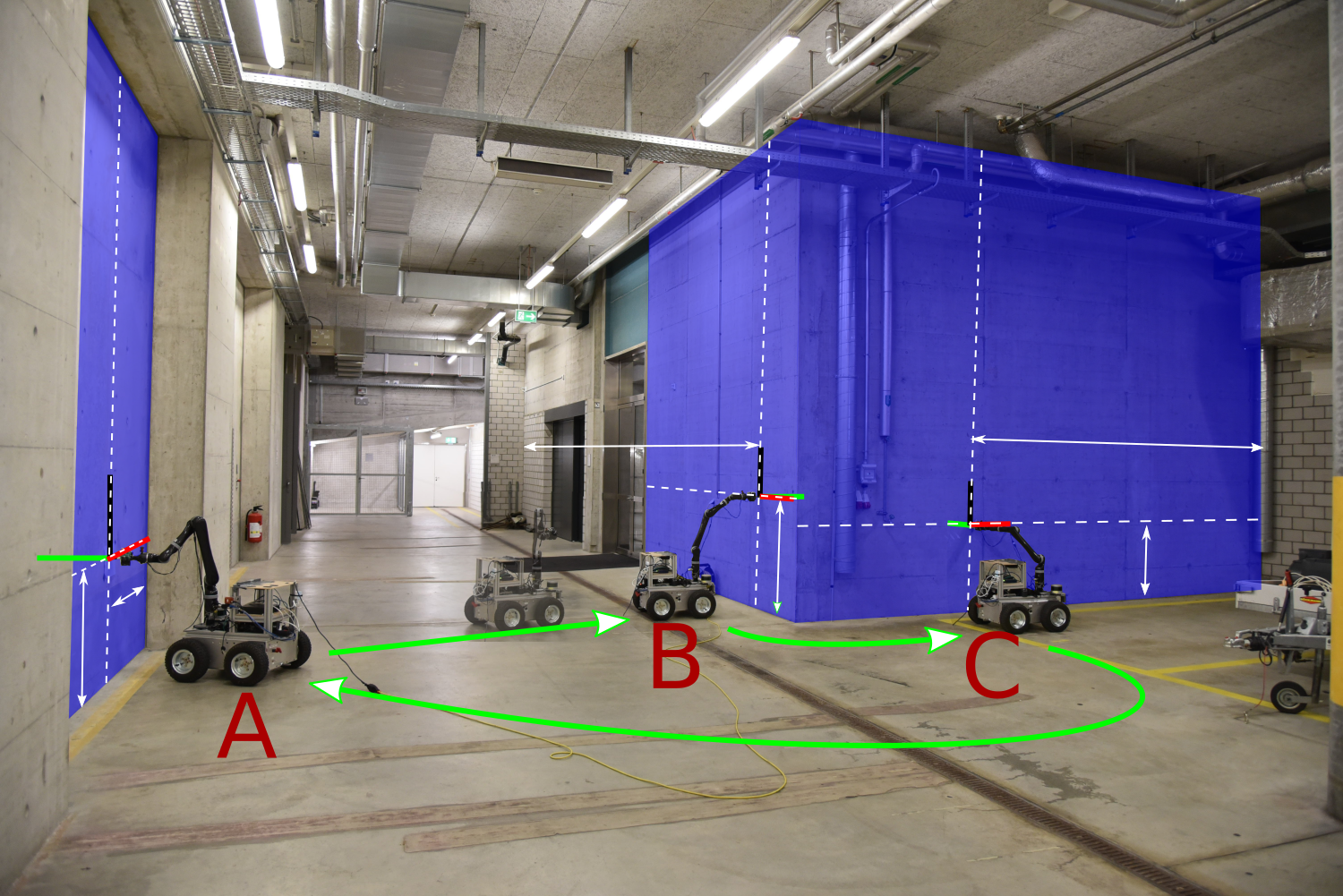}
    \caption{The proposed system enables high-accuracy interaction of a mobile robot with the environment using only on-board sensing. Here, three task locations are highlighted between which a robot traversed in our experiments yielding sub-\SI{}{cm} errors.}
    \vspace{-1.5em}
    \label{fig:teaser}
\end{figure}
Moreover, as every building project is unique, a key challenge is to find strategies to cope with the inaccuracies of the building materials and components~\cite{ardiny2015autonomous}. Robot world representations are usually referenced to one globally consistent map, or a number of sub-maps. Yet, the execution of building tasks primarily depends on the as-built status of existing installations, i.e., a set of local references and dependencies.
Finally, motion planning and execution on construction sites is also a challenging problem. The typical kinematic redundancy of a mobile manipulator needs to be leveraged appropriately in order to coordinate locomotion and manipulation, while taking into account the physical constraints of the environment and uncertainties in the sensing~\cite{schwartz2017robotic}.

While existing solutions rely on external sensing systems, or augmentation of the environment with specific markers to achieve the accuracy required in construction tasks, this work presents an autonomous mobile robotic solution for highly accurate building task execution, solely using on-board sensing.
The system is based on the comprehensive integration of software solutions on a mobile platform to perform the workflow from extracting tasks from building plans to task execution. It automatically derives construction tasks from a 3D building model, enables the robot to localize within the model using LiDAR sensors, uses local references for highly-accurate localization, plans trajectories, dynamically avoids obstacles, and implements a whole-body motion planning regime distributing the movement between the mobile base and robotic manipulator.
We evaluate the robot's performance towards discrete building tasks, e.g. drilling, under realistic conditions.
This work presents the following contributions:
\begin{itemize}
    \item A novel optimization-based method for high-accuracy robot localization using a set of plane intersections between laser distance sensor measurements and a 3D building model.
    \item LiDAR-based localization within a 3D building model using 3D scan matching.
    \item Whole-body model-predictive planning and control for end-effector pose tracking of a wheeled mobile manipulator.
    \item Experimental validation of the proposed system on a prototype robot in a realistic construction-like environment.
\end{itemize}
The remainder of this paper is organized as follows. In Section~\ref{sec:related_work}, we review the related work on construction robotics, and related sensing solutions. Section~\ref{sec:system} introduces the sensing and control system developed within this paper followed by an experimental evaluation of the system performance in Section~\ref{sec:experiments}, and concluding remarks in Section~\ref{sec:conclusions}.
\section{Related Work}
\label{sec:related_work}
Construction robotics and digital fabrication are emerging topics in the robotics research community~\cite{ardiny2015autonomous}.
Existing solutions towards construction automation range from fixed automation solutions~\cite{bosscher2007cable} over innovative systems, e.g. using flying robots~\cite{willmann2012aerial}, to multi-purpose mobile ground robotic systems~\cite{willmann2012aerial, helm2012mobile,sandy2016autonomous}.
The proposed system is within the last category.
Multi-purpose ground-robotic solutions are converging on typical design aspects, consisting of robotic mobile bases, manipulators, task-specific end-effectors, and sensing suite capabilities~\cite{helm2012mobile, zhang2018large, feng2015vision, keating2017toward, sandy2016autonomous, dorfler2016mobile}.
Localizing such robots within the construction environment can be a difficult task, as perceptual conditions are challenging for cameras, and environments cannot easily be augmented with external localization markers or beacons, as used in~\cite{feng2015vision, zhang2018large}.
To enable automatic task generation from digital building models and enable robots to autonomously travel between them, it is further required to not only identify the working location, but also localize within the building model of the environment.
\cite{luo2010autonomous} match corner-, and door plate-detection with cameras against 2D maps extracted from 2D floor plans.
\cite{boniardi2017robust,boniardi2019pose} localize within floor plans abstracted to 2D grid maps using G-ICP. They further robustify their localization routine using a pose-graph SLAM system that enables the robot to account for largely cluttered environments as compared to the clean floor plan map.
A solution using an initial reference scan of the environment using 3D LiDAR scans is presented in \cite{dorfler2016mobile}.
Similarily, \cite{rowekamper2012position} and \cite{vasiljevic2016high} use scan matching and MCL in 2D using 2D LiDAR sensors.
However, no domain shift in localization between environment model and LiDAR scans is performed, as the work location is inserted into the environment represented by the reference scan.
Hence, the accuracy is limited by the reference scanner characteristics.
This paper, however, presents a solution facilitating 3D building models for localization of 3D LiDAR scans.
The routine performs ICP between LiDAR scans and the point-cloud sampled from a 3D CAD triangular mesh model.

Another key challenge for construction robots is high-accuracy end-effector positioning~\cite{ardiny2015autonomous}.
This requires highly-accurate knowledge about the end-effector position, e.g., via on-board sensing~\cite{feng2015vision, zhang2018large,sandy2016autonomous,dorfler2016mobile} or external positioning systems~\cite{willmann2012aerial, keating2017toward}.
Additionally, the positioning of the end-effector and task execution require either stiff structure~\cite{sandy2016autonomous} or high control frequency~\cite{keating2017toward}.
Using 2D LiDAR sensors, it is possible to perform both the localization in the previously recorded model and high-accuracy localization using the same scan-matching routine~\cite{dorfler2016mobile, rowekamper2012position,vasiljevic2016high}.
Yet, high-accuracy on-board localization can also be realized with local visual markers and their detection using cameras~\cite{feng2015vision, zhang2018large}, reaching accuracies well below \SI{1}{\centi\meter}.
\cite{rowekamper2012position} and \cite{vasiljevic2016high} report localization accuracy using scan matching and MCL below \SI{1}{\centi\meter} in 2D using 2D LiDAR sensors in static environments, with an accuracy only mildly suffering under changed scenarios.
\cite{sandy2016autonomous} use 3D LiDAR scans to precisely localize in 3D against the known model of a work-piece (in contrast to localizing within the environment), given a localization prior, achieving an average accuracy of $\pm$5\SI{}{\milli\meter}.

In this work, we extend this concept facilitating the prior from 3D LiDAR localization, and inexpensive laser distance measurement sensors to perform local localization at \SI{}{\milli\meter} scale via measurement intersections with the known building model.

Many of the previous works have considered executing construction tasks with mobile manipulators. Such robots must locate tools mounted on the manipulator's end-effector with high accuracy. In order to do so, robot motions are typically decomposed into two sequential steps of first repositioning the robot's base, then moving the arm to bring the tool to the desired location~\cite{dorfler2016mobile,sandy2016autonomous,keating2017toward,zhang2018large}. To the best of our knowledge, this is the first work that adaptively plans and tracks whole-body motions for the high-accuracy building task execution. In doing so, our system can smoothly position its tool from optimal body postures.
\section{Sensing and control system}
\label{sec:system}
In this section, we present the sensing and control system for high-accuracy mobile manipulation.
The system covers the full range of functionalities to enable a mobile robot to receive tasks from a building model, autonomously localize within it, reach indicated task locations, and position an end-effector tool with \SI{}{\milli\meter}-accuracy.
The system solely relies on on-board sensing and requires no human intervention, however allows human intervention if necessary.
Key software components are illustrated in Fig.~\ref{fig:overview}.

A central role is given to the high-level manager that allocates tasks received from a building task interface to various software components.
We consider building tasks which require the high-accuracy positioning of an end-effector-mounted tool at discrete locations in place, such as drilling, anchoring, and pick-and-place tasks. Expanding the system to support tasks which require either control of the tool's interaction force with the environment (e.g., chiseling) or the high-accuracy tracking of the tool along a continuous trajectory (e.g., plastering) is planned for future work.
The mode of the system is switched depending on the distance from the current to the desired tool pose.
The state is estimated via a moving horizon estimator (MHE), fusing LiDAR localization updates, IMU measurements, and wheel odometry~\cite{sandy2019confusion}.
The estimated state is used in feedback by a whole-body motion controller which tracks base and end-effector reference trajectories. These references are generated by a Model Predictive Controller (MPC) strategy running at $\sim$100 Hz. When the tool is far from the target pose, a stochastic navigation planner is used to find a collision-free trajectory to bring the robot near the target location. This base trajectory is used to seed the MPC. Once sufficiently close, the MPC iteratively plans and tracks a whole-body trajectory to complete the task.
Finally, to reach highest accuracy, an additional localization routine is performed near task locations using the localization of the state estimator as an initial guess, refined with an independent sensor set.
\pgfdeclarelayer{background}
\pgfdeclarelayer{foreground}
\pgfsetlayers{background,main,foreground}
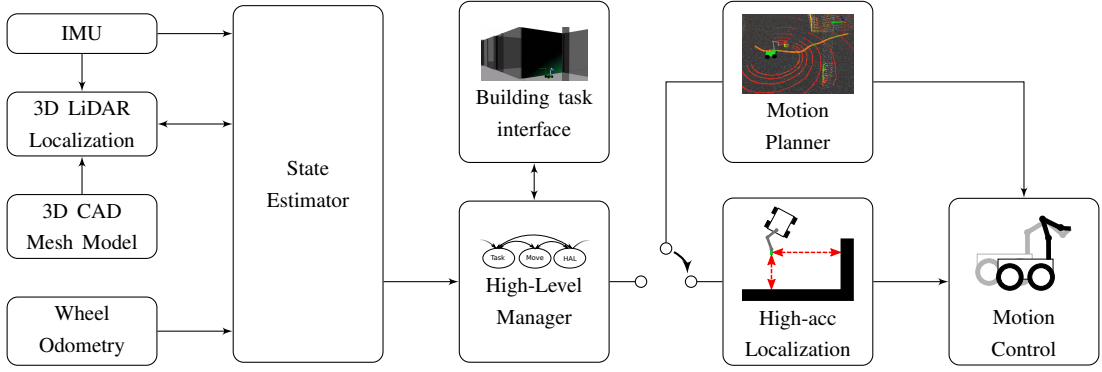
\begin{figure*}
\centering
\begin{tikzpicture}[block/.style   ={rectangle, draw, text width=5em, text centered, rounded corners, minimum height=3em, minimum height=6.1em}, blocki/.style   ={rectangle, draw, text width=5em, text centered, rounded corners, minimum height=3em, minimum height=1.5em}, blockj/.style   ={rectangle, draw, text width=5em, text centered, rounded corners, minimum height=3em, minimum height=13.5em},
node distance=2.1cm, auto, >=latex', circuit ee IEC]

	\node [blocki, align=center] (imu) {\footnotesize IMU};
	\node [blockj, align=center, right = 1cm of imu, yshift=-2.0cm] (state) {\footnotesize State\\ \footnotesize Estimator};
	\node [block, align=center, right = 1cm of state, yshift=-1.3cm] (hlm) {\includegraphics[width=4.2em]{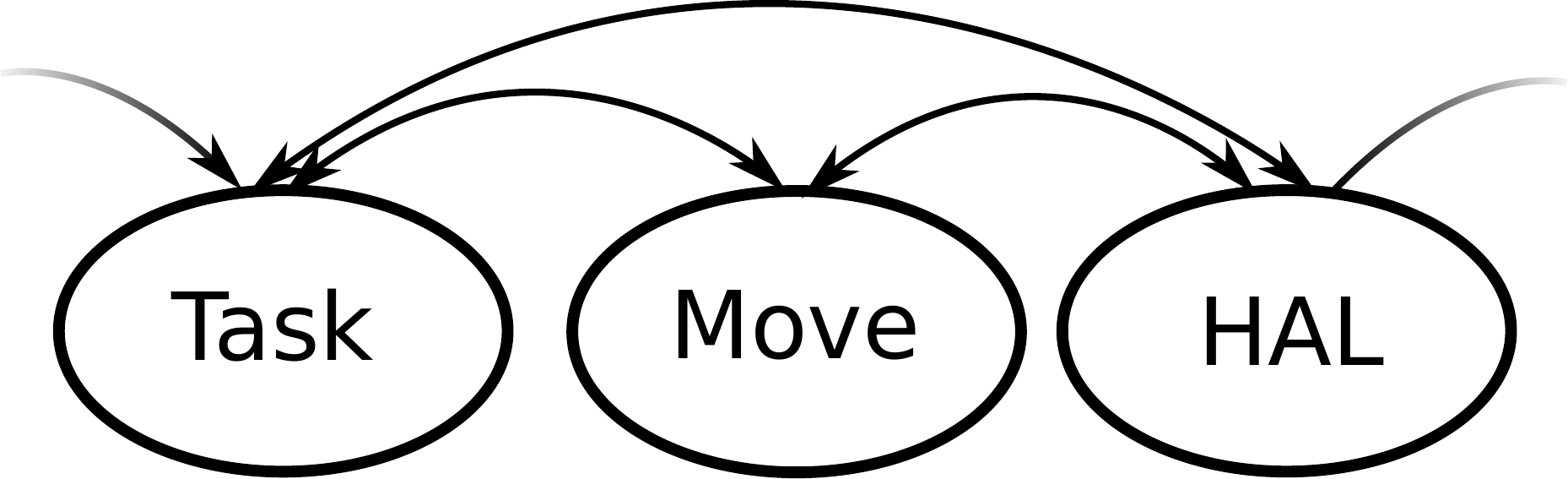} \\ \footnotesize High-Level\\ \footnotesize Manager};
	\node [block, align=center, above = 0.5cm of hlm] (rhino) {\includegraphics[width=4.0em]{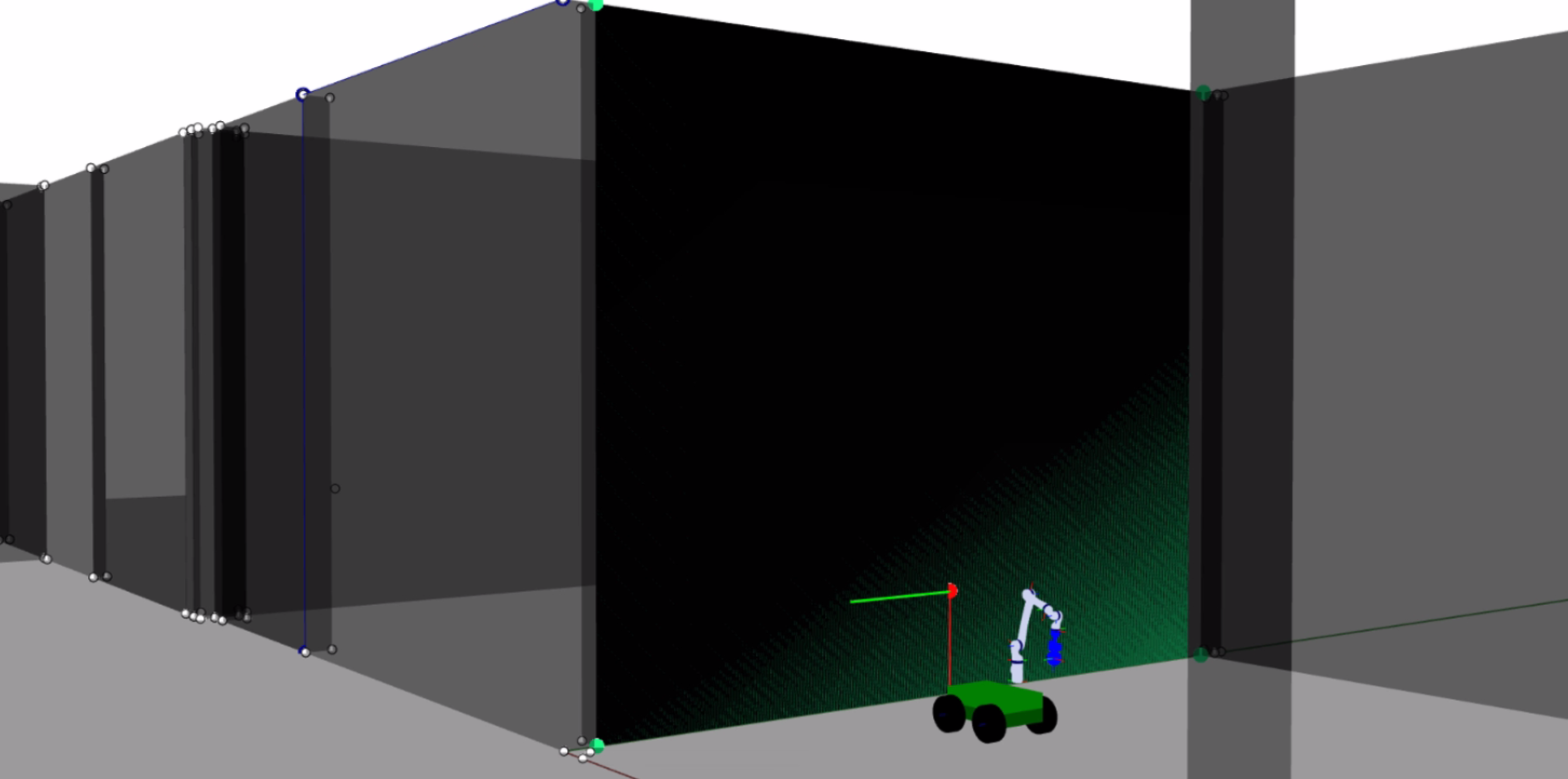} \\ \footnotesize Building task \\ \footnotesize interface};
	\node [block, align=center, right = 1.5cm of rhino] (planner) {\includegraphics[width=4.2em]{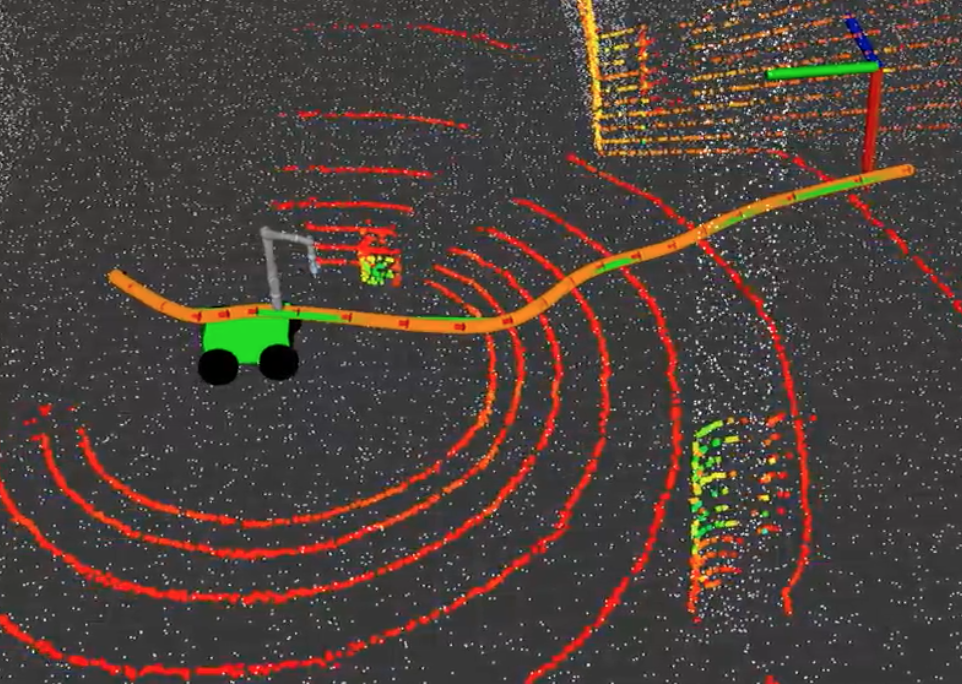} \\ \footnotesize Motion \\ \footnotesize Planner};    
	\node [block, align=center, right = 1.5cm of hlm] (hal) {\includegraphics[width=4.2em]{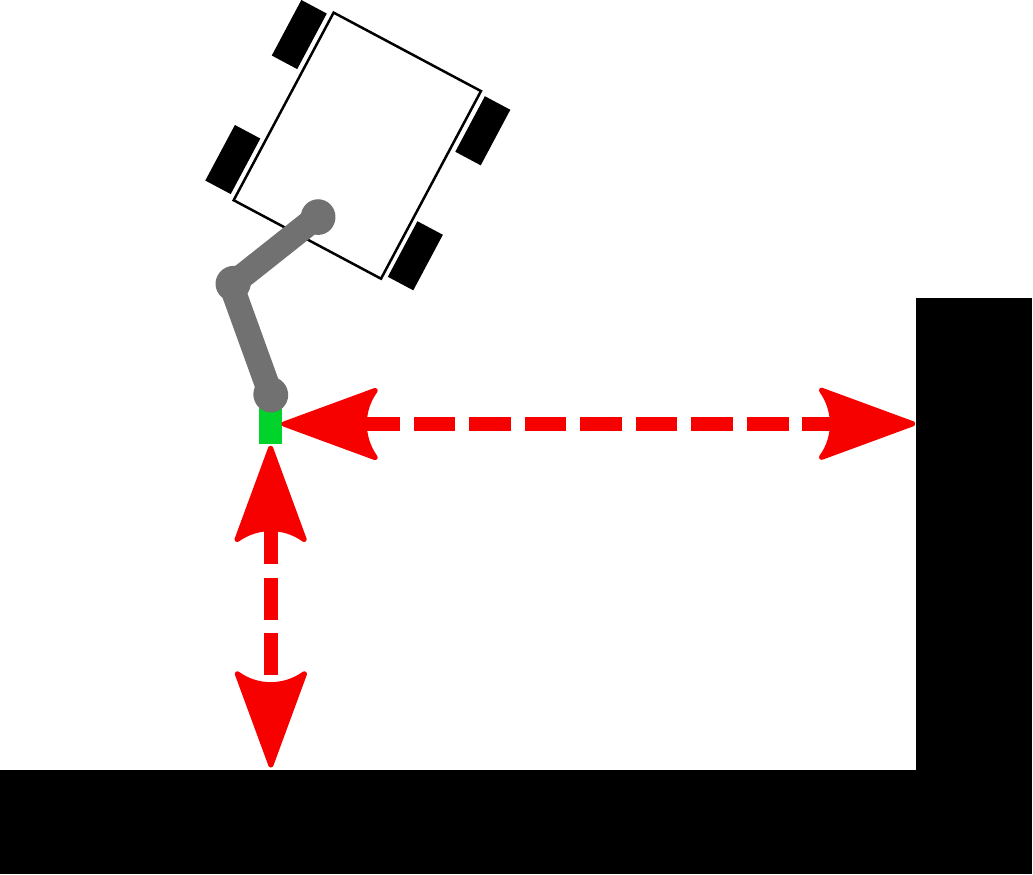} \\ \footnotesize High-acc \\ \footnotesize Localization};
	\node [block, align=center, right = 1.0cm of hal] (task) {\includegraphics[width=4.2em]{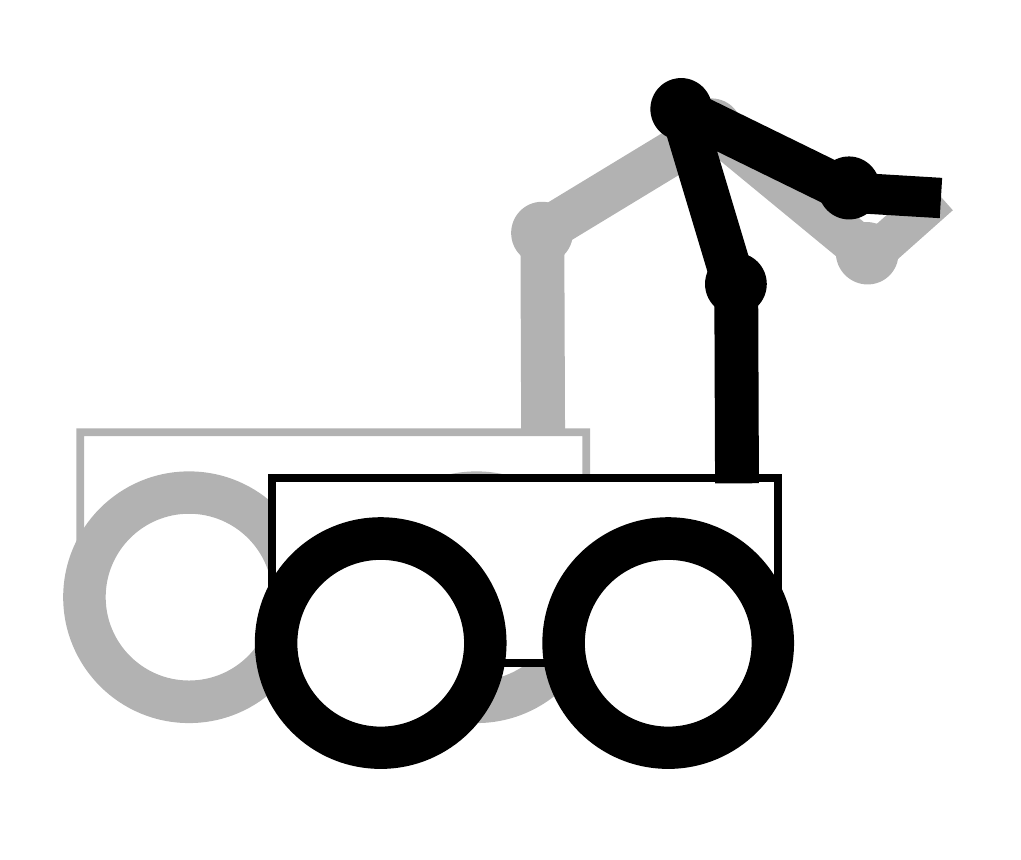} \\ \footnotesize Motion \\ \footnotesize Control};
	\node [blocki, align=center, below = 0.5cm of imu] (lidarslam) {\footnotesize 3D LiDAR \\ \footnotesize Localization};
	\node [blocki, align=center, below = 0.5cm of lidarslam] (cad) {\footnotesize 3D CAD Mesh Model};
	\node [blocki, align=center, below = 0.5cm of cad] (odometry) {\footnotesize Wheel \\ \footnotesize Odometry};

\draw [<->] (rhino) -- node {} (hlm);
\draw[<->] (lidarslam.east)  -- (lidarslam-|state.west);
\draw [<-] (hlm.west) -- node {} (hlm-|state.east);
\draw [->] (planner) -| node {} (task);
\draw [->] (hal) -- node {} (task);
\draw [->] (imu) -- node {} (lidarslam);
\draw[->] (cad)  -- (lidarslam);
\draw[->] (imu.east)  -- (imu-|state.west);
\draw[->] (odometry.east)  -- (odometry-|state.west);
\draw[-o] (hlm.east) -- ++(0.5cm,0) coordinate(r1){}; 
\draw[-o] (planner.west) -- ++(-0.75cm,0) -- ++(0,-2.3cm) coordinate(b1){};
\draw[-o] (hal.west) -- ++(-0.5cm,0) coordinate(a1){};
\draw[->, thick] ($(b1)+(3pt,1pt)$) to [bend left ]($(a1)+(1pt,3pt)$);
\end{tikzpicture}
\caption{Software system overview of the mobile construction robot: The building task interface provides task locations within the building model of the environment to the high-level robot manager. The high-level manager requests the current robot state fused from ICP-based localization within the building model, IMU measurements, and wheel odometry. Depending on the robot state, the manager switches between planning robot base trajectories to the task location, and a \ac{HAL} procedure depending on the distance to the target. Once the robot is close to the target, a \ac{HAL} procedure using laser distance sensors is triggered. The commands are executed using a whole body motion controller.}
\vspace{-1.5em}
\label{fig:overview}
\end{figure*}
\subsection{State Estimation}
State estimation of the robot's base-pose $\boldsymbol{x}_{t_n}$ at time $t_n$ is performed using the \emph{ConFusion}~\cite{sandy2019confusion} moving horizon estimator, minimizing an error over the various inputs in the least-square sense.
Individual errors $\boldsymbol{e}_j$ are calculated using measurements $\boldsymbol{z}_j$ connecting to one or more states $\boldsymbol{x}_{t_{i-j}}$, and static parameters $\boldsymbol{s}$, such as calibrations within the function $\boldsymbol{h}$, i.e.,
\begin{equation}
    \boldsymbol{e}_j = \boldsymbol{h}(\boldsymbol{x}_{t_{i-j}:t_{i}},\boldsymbol{s}, \boldsymbol{z}_j)
\end{equation}
We thus optimize for the states $\boldsymbol{x}_{t_o:t_n}$, and static parameters $\boldsymbol{s}$, forming an optimization problem over $n$ state instances, i.e.,
\begin{equation}
    \{\boldsymbol{x}_{t_o:t_n}, \boldsymbol{s}\} = \argmin_{\boldsymbol{x}_{t_o:t_n}, \boldsymbol{s}}\sum_{i=0}^n{\sum_{\boldsymbol{z}_j\in \boldsymbol{Z}_{t_{i-j}:t_i}}{\norm{\boldsymbol{e}_j}^2}}
\end{equation}
Different sensor modalities contribute error terms to the optimizer.
Our system implements two types of state constraints, i.e., pose constraints ($j=0$), and relative pose constraints ($j=1$).
\subsubsection{3D LiDAR Localization}
We facilitate 3D LiDAR to localize the robot within the building model of the environment.
The map $\boldsymbol{\mathcal{M}}$ is represented by a sampled point-cloud from the 3D CAD traingle mesh model $\boldsymbol{M}$. Motion compensated 3D LiDAR scans are continuously matched against the model using ICP. 
ICP performs a separate minimization, of the error between new LiDAR points $\boldsymbol{p}_k\in{\boldsymbol{\mathcal{L}_{t_n}}}$, and matched map points $\boldsymbol{q}_k\in{\boldsymbol{\mathcal{M}}}$ with associated surface normals $\boldsymbol{n}_k$, facilitating a point-to-plane error metric, i.e.,
\begin{equation}
\boldsymbol{\hat{x}}_{t_n}^{ICP} = \argmin_{\boldsymbol{\hat{y}}_{t_n}}{ \sum_{k=0}^K{\norm{(\boldsymbol{p}_{k}-\boldsymbol{q}_k)\boldsymbol{n}_k}}}
\end{equation}
The resulting robot base pose updates $\boldsymbol{\hat{x}}_{t_n}^{ICP}\subseteq{{\boldsymbol{x}}_{t_n}}$, are then sent to the state estimator, assuming that the pose updates are corrupted by a constant normally-distributed measurement noise.
While we focus on localization within the building model in this work, the localization module is also able to build a map by incrementally fusing consecutively matched scans, and initialized at the starting location of the robot.
An approximate initial guess within the environment is given by the known starting location.
A global localization scheme is outside the scope of this paper.
\subsubsection{Wheel Odometry}
Rotary encoders mounted on a robot's wheels are used to measure the angular velocity of the wheels. Considering a simple differential drive kinematic model, these measurements provide a process constraint on the evolution of the estimated base pose. The modeled confidence in this constraint is much lower when the wheels are turning than when they are stationary, since the simple kinematic model does not capture wheel slippage and non-flat ground profiles. This constraint therefore mostly helps enforce that the base remains stationary when the wheels are not turning.
\subsubsection{IMU}
Angular velocity and linear acceleration measurements from an inertial measurement unit (IMU) provide an additional process constraint on the evolution of the base pose. The linear velocity, gyroscope bias, and accelerometer biases are also estimated online as is typically done for inertial sensor fusion. The full model and conventions used are described in a previous work~\cite{sandy2017dynamically}. Because of their high rate and low latency, the IMU measurements are additionally used to compensate for the latency of the LiDAR pose updates. When estimates are received by the robot controller, they are forward propagated through the more recent IMU measurements up to the current controller time. Since the underlying MHE problem is solved using non-linear optimization, as opposed to more commonly used filter-based methods, fusion of the multiple process measurements does not require augmenting the state with additional terms, as in~\cite{kubelka2015robust}.
\subsection{High Accuracy Localization (HAL)} 
Both, the measuring accuracy of the 3D LiDAR and discrepancies between the building model and the as-built conditions of the environment, limit the accuracy of the state estimate.
For building task execution, the system therefore implements a \ac{HAL} strategy against the task-relevant references, i.e., walls around the task space.
 Initialized with the state estimator's current pose, we localize the static robot base directly in the building model, represented as a 3D triangular mesh, using high accuracy laser distance sensors mounted at the end-effector.\\
This is done by optimizing the error between expected and measured distances of the laser sensors for several end-effector poses $\boldsymbol{y}_{t_n}^{(i)}$ (6 in our experiment, for ease of notation, the index $(i)$ will be omitted in the following). At each iteration, we find the planes against which the laser distance sensors are measuring using ray-tracing in the building model from the current estimate of the sensor poses at the end-effector. Given the plane support $\boldsymbol{t}_p$ and the plane normal $\boldsymbol{r}_p$ of the intersected triangle in the mesh, we perform an optimization over the locally differentiable error function $e_{t_n}^{HAL}(\boldsymbol{y}_{t_n})$:

\begin{align*}
    e_{t_n}^{HAL}(\boldsymbol{y}_{t_n}) = {\sum \limits_i c\left(\norm{z_i - \frac{\left(\boldsymbol{t}_p(\boldsymbol{y}_{t_n}^{ }) - \boldsymbol{y}_{t_n}^t \right)\boldsymbol{r}_p(\boldsymbol{y}_{t_n}^{ })}{\boldsymbol{y}_{t_n}^R \boldsymbol{r}_p(\boldsymbol{y}_{t_n}^{ }) }} \right)}
\end{align*}

\begin{align*}
    \boldsymbol{y}_{t_n}^{HAL} = \argmin_{\boldsymbol{y}_{t_n}}{e_{t_n}^{HAL}(\boldsymbol{y}_{t_n})}
\end{align*}

with $\boldsymbol{y}_{t_n}^t$, and $\boldsymbol{y}_{t_n}^R$ the translational, and rotational components of $\boldsymbol{y}_{t_n}$ respectively.
To account for problematic measurements against clutter or windows and errors in plane retrieval due to initial pose uncertainty, we use a Cauchy robust cost function $c$. The relative transformations between the end-effector poses are assumed to be known by the manipulator's joint encoder readings. The procedure is formalized in Algorithm~\ref{algorithm}.

During the development of the system, we found that the estimate of the orientation with this routine was not as reliable and accurate as the initial estimate from aligning the much richer LiDAR scan with the building model. We therefore keep the orientation component fixed to the initial estimate and only optimize over the position. Doing so further increases the precision of the corrected position.

These local pose corrections $\boldsymbol{y}_{t_n}^{HAL}$ are only used for the local task execution, and not part of the general state estimation framework.
\setlength{\textfloatsep}{2pt}%
\begin{algorithm}
    \SetKwInOut{Input}{Input}
    \SetKwInOut{Output}{Output}

    \underline{function localize} $(\boldsymbol{y}_{t_n},\boldsymbol{S},m,\boldsymbol{M})$\;
    \Input{Initial guess for pose of end-effector $\boldsymbol{y}_{t_n}$, extrinsic calibration of sensors $\boldsymbol{S}$, iterations $m$, and 3D triangle mesh model $\boldsymbol{M}$}
    \Output{corrected pose of end-effector $\boldsymbol{y}_{t_n}^{HAL}$}
    $e_{t_n}^{HAL}(\boldsymbol{y}_{t_n})=0$\\
    $\boldsymbol{y}_{t_n}^{(i)}\gets \boldsymbol{y}_{t_n}$\\
    \For{$i=1:n$}
      {
        $\boldsymbol{y}_{t_n}^{(i)} \gets$ moveSensorHead($\boldsymbol{y}_{t_n}^{(i)}$);\\
        \For{$j=1:\text{num\_sensors}$}
          {
            $\boldsymbol{y}_{t_n}^{(j)}\gets$ retrievePose($\boldsymbol{y}_{t_n}^{(i)}, \boldsymbol{S}$)\\
            $z_j\gets$measureDistance()\\
            $\boldsymbol{t}_p, \boldsymbol{r}_p\gets$ raytrace($\boldsymbol{y}_{t_n}^{(j)},\boldsymbol{M}$)\\
            $e_j\gets c\norm{z_j - \frac{\left(\boldsymbol{t}_p(\boldsymbol{y}_{t_n}^{(j)}) - \boldsymbol{y}_{t_n}^{(j),t} \right)\boldsymbol{r}_p(\boldsymbol{y}_{t_n}^{(j)})}{\boldsymbol{y}_{t_n}^{(j),R} \boldsymbol{r}_p(\boldsymbol{p}_{j}) }}$\\
            $e_{t_n}^{HAL}(\boldsymbol{y}_{t_n})$ += $e_j$\\
          }
      }
    $\boldsymbol{y}_{t_n}^{HAL}\gets \argmin_{\boldsymbol{y}_{t_n}}{e_{t_n}^{HAL}(\boldsymbol{y}_{t_n})}$\\
      {
        return $\boldsymbol{y}_{t_n}^{HAL}$;
      }
    \caption{Proposed algorithm for \ac{HAL}}
    \label{algorithm}
\end{algorithm}
\subsection{Building Task Interface}
The building task interface bridges the gap between the design and planning environment containing the complex building information, and the robotic system. On one side, this serves as a suitable interface for users of the robotic system facilitating a building process that allows simple task-level commanding. On the other side, it seamlessly integrates the workflow between building construction and the robotic system for executing building tasks. 

 A 3D building model is built within the open-source, Python-based computational framework COMPAS\footnote{\href{ https://compas-dev.github.io/}{ https://compas-dev.github.io/}} and implemented in Rhino Grasshopper to simulate robot state and task status in relation to the building model. The kinematic model of the robot is visualized using the \texttt{compas\_fab} package of the COMPAS framework. Communication between high-level manager and building environment is established using robotic middle-ware \texttt{roslibpy}\footnote{\href{ https://roslibpy.readthedocs.io/}{ https://roslibpy.readthedocs.io}}. This interface between building task and high-level manager enables automatic task generation, monitoring, and user intervention. Building tasks defined within the building model are thus converted to end-effector poses represented in the reference frame of the robot and sent to the high-level manager. 

While the current system uses feed-forward task generation from the building model, the system is designed to receive updates from the high level-manager for the as-built status of the construction environment to update the building model and robot tasks. Furthermore, planned for future work, it will enable leveraging task-specific references, such as relative distances to selected building components, and relative tolerances between consecutive building tasks.

\subsection{Motion Planning} \label{ssec:motion_planning}
The motion planning module generates robot state reference trajectories to bring the end-effector to the target pose commanded by the building task interface. A typical task consists of: globally navigating the wheeled base through the workspace to reach the task execution area, jointly controlling base and arm motion to get the end-effector near the commanded pose, executing the \ac{HAL} routine based on the laser distance sensors on the end-effector, then finally fine adjusting the end-effector to the commanded pose.

The planning module is made up of three components in order to address these subtasks. When the robot is far away from its target, a stochastic path planner computes planar base trajectories for navigation through the workspace towards the target. The planner takes into account both prior-knowledge of the building structure as well as LiDAR data, which enables re-planning for dynamic obstacle avoidance. An MPC module is responsible for generating whole-body motion references. As an input it takes either the base motion plan from the navigation planner, or, when the robot is close enough to its target, a desired end-effector pose. In both cases a continuous optimal whole-body motion plan, comprised of both base and arm motion, is computed to track the desired references. In order to accurately bring the end-effector to the desired poses required for the \ac{HAL} routine and execution of the building task, the base is assumed to be stationary and a pose interpolation approach is used, combined with an iterative inverse kinematics algorithm, in order to compute desired joint positions. These three components of the planning module are described in more detail below.

\subsubsection{Base Navigation Planning}
The objective of the base navigation planner is to drive the robot close to the area where the tasks need to be executed. Collision-free trajectories are computed in the workspace by running the RRT* algorithm \cite{Karaman2011sampling}, while checking for collisions by means of Octomap \cite{hornung13auro}. The planner adopts an \textit{optimistic} behavior by assuming the unknown space to be free. The initial occupancy map of the environment is generated using the point-cloud sampled from the 3D CAD triangle mesh model of the robot's environment and directly used for planning. The initial map is then updated incrementally using the LiDAR scans acquired during navigation, in order to identify obstacles which are not captured by the building model or to clear free areas which were previously considered as occupied. In case the building model is not accessible, the occupancy map is built incrementally from scratch during the navigation. Collision checks are performed along the planned trajectory every time the occupancy map is updated considering a rectangular bounding box around the robot. This allows for fast reactions to obstacles that appear and may block the path during navigation. Every time a potential collision is identified along the current planned path, the planner is triggered and a new trajectory that considers the updated map of the environment is computed.

\subsubsection{Whole-Body MPC}
The MPC module, implemented with the OCS2 framework~\cite{farshidian2017MPC}, uses a kinematic plant model of the robot. The robot state consists of the base position and orientation in the horizontal plane and the arm joint angles $\boldsymbol{x} = (x_b, y_b, \theta{}_b, q_0, \dots , q_5)^T$. The module's outputs are the base linear velocity, the steering rate in the horizontal plane and the arm joint angular velocities $\boldsymbol{u}=(v_b, \dot{\theta}_b, \dot{q}_0, \dots , \dot{q}_5)^T$. The output is forwarded to the inverse dynamics tracking controller.

Two different types of references are used. To track the base motion plan, the references $\boldsymbol{x}_j$ are the base position and orientation and default joint angles for the arm. For generating whole body motion trajectories the desired end-effector pose $\boldsymbol{x}_{ee}$ is used as a reference.
The controller tracks both references with different cost functions. A unified cost function is set up as the sum over both individual costs to track both references with a single MPC strategy:
\begin{align}
    J(\boldsymbol{x}, \boldsymbol{\Tilde{x}},\boldsymbol{u}) =& \: \int_{t_0}^{t_0+T} (1-\alpha) C_{j} + \alpha C_{ee} + \boldsymbol{u}^T \boldsymbol{R} \boldsymbol{u} \: d\tau \label{eq::cost_sum} \nonumber \\
    & + \alpha \boldsymbol{\Phi}_{ee}
\end{align}
\begin{align}
    C_{j} =&  (\boldsymbol{x} - \boldsymbol{\Tilde{x}}_{j})^T \boldsymbol{Q}_{j} (\boldsymbol{x} - \boldsymbol{\Tilde{x}}_{j}) \label{eq::cost_joint} \\
    C_{ee} =& (\boldsymbol{\xi}(\boldsymbol{x}) \boxminus \boldsymbol{\Tilde{x}}_{ee})^T \boldsymbol{Q}_{ee} (\boldsymbol{\xi}(\boldsymbol{x}) \boxminus \boldsymbol{\Tilde{x}}_{ee}) \nonumber \\ & + (\boldsymbol{x} - \boldsymbol{\Tilde{x}}_{d})^T \boldsymbol{Q}_{d} (\boldsymbol{x} - \boldsymbol{\Tilde{x}}_{d}) \label{eq::cost_ee} \\
    \boldsymbol{\Phi}_{ee} =& (\boldsymbol{\xi}(\boldsymbol{x}(T)) \boxminus \boldsymbol{\Tilde{x}}_{ee}(T))^T \boldsymbol{Q_{ee, \Phi}} (\boldsymbol{\xi}(\boldsymbol{x}(T)) \boxminus \boldsymbol{\Tilde{x}}_{ee}(T))
\end{align}
The parameter $\alpha \in \{0,1\}$ determines which cost is active and negotiates a smooth transition between the two operating modes.
The cost $C_j$ penalizes deviations of the current robot state $\boldsymbol{x}(t)$ to the base motion plan $\boldsymbol{\Tilde{x}}_j$ with $\boldsymbol{Q}_j=\boldmath{\mathbb{1}}$. $C_{ee}$ penalizes end-effector tracking errors with $\boldsymbol{Q}_{ee} = \boldsymbol{\mathbb{1}}$ and an additional weaker cost encourages maintaining a default arm configuration with $\boldsymbol{Q}_{d} = 0.01 \: diag(0, 0, 0, 3, 10, 10, 0, 5, 0)$. The small weight on the default configuration helps to avoid self-collisions and collisions of the base with the wall. $\boldsymbol{R}=\mathbb{1}$ puts costs on the control inputs $\boldsymbol{u}$.
An additional terminal cost $\boldsymbol{\Phi}_{ee}$ with $\boldsymbol{Q}_{ee, T} = 10 \: \boldsymbol{\mathbb{1}}$ penalizes deviations from the desired end-effector pose at the end of the MPC horizon.
The transformation function $\boldsymbol{\xi}$ from the robot state $\boldsymbol{x}(t)$ to the end-effector pose is autogenerated by the RobCoGen library \cite{frigerio2016robcogen}. The code is templated over the scalar type such that the automatic differentiation toolbox CppAd \cite{bell2012cppad}, used in OCS2, can generate derivatives of the cost function to solve the MPC problem \cite{giftthaler2017ad}. With the distance measure~$\boxminus$, we subtract the positions of the two poses and use a quaternion distance measure for the orientations \cite{siciliano_robotics:_2009}.
The MPC is solved over a time horizon of $T = \SI{1}{\second}$. If the desired target cannot be reached within the current time horizon, the references are interpolated and for each roll-out a reference is chosen such that it can be reached in the given time, assuming some nominal speed.

\subsubsection{End-effector Pose Interpolation and Inverse Kinematics} \label{sssec:pose_interpolation}
For the situations in which the base is stationary, smooth motion reference paths between desired end-effector poses are obtained using pose interpolation. This is done by defining a smooth interpolation profile with zero velocity and acceleration at the start and end, and applying that to linearly interpolate between position vectors and using spherical linear interpolation for the orientation quaternions. To compute the arm joint positions for the desired end-effector poses, we use an iterative inverse kinematics scheme as described in \cite{goldenberg1985complete}.

\subsection{Whole-Body Motion Control}
The motion references generated by the components of the planning module (described in Sec.~\ref{ssec:motion_planning}) are tracked by a motion controller. Two separate motion controllers are used for tracking the whole-body MPC motion references and for precisely tracking end-effector pose references when the base is stationary.

During navigation and whole-body end-effector motion tracking, the motion references come from the MPC in the form of velocities for the mobile base in the horizontal plane and the six arm joints. The velocity references for the wheeled base are translated into feed-forward wheel velocity references $\boldsymbol{v}^{ff}_{wheel}$ using a two-wheeled differential drive model of the robot. Because the actual base has four wheels and depends on a skid-steering mechanism which introduces a significant amount of wheel slip, an additional term $\boldsymbol{v}^{i}_{wheel}$ is added based on the integrated error between the desired and measured linear and angular base velocities. The final wheel velocity references are computed as $\boldsymbol{v}^{ref}_{wheel} = \boldsymbol{v}^{ff}_{wheel} + \boldsymbol{v}^{i}_{wheel}$ and are tracked by the base motor controller. For the arm joints, the reference velocities are numerically integrated to obtain position references, and numerically differentiated and low-pass filtered to obtain acceleration references. These three levels of motion references are then tracked using torque-control. This allows for compliant and safe interaction when the arm comes in contact with humans or its environment. The required joint torques are computed using a numerical inverse dynamics controller as proposed in \cite{bellicoso2017dynamic}. Computation of the joint-torque references takes into account the system dynamics, desired joint motions, and the actuator torque limits. To compensate for unmodeled dynamics, such as joint friction, we update these torque references with additional desired torques based directly on the joint position and velocity errors. These joint torque references are then tracked by the arm's internal controller.

When executing the \ac{HAL} scanning routine and moving to the final task pose the arm actuators are switched to position-control mode. In this situation, the joint position references from the inverse kinematics solver (Sec.~\ref{sssec:pose_interpolation}) are commanded directly.

\section{Experiments} \label{sec:experiments}
We evaluate the system on a prototype robotic platform within a realistic construction environment.
\subsection{Hardware Setup}
\begin{figure}
    \centering
    \includegraphics[width = 0.9\columnwidth]{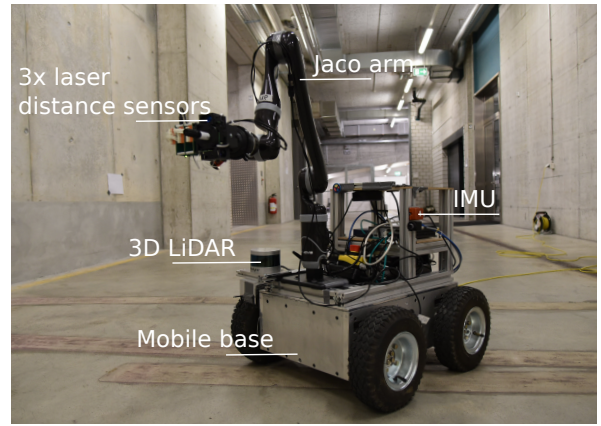}
    \caption{The mobile construction robot research platform \emph{Waco} used in this work, equipped with 6 \ac{DoF} arm, and sensors.}
    \label{fig:waco}
\end{figure}

The robot used for the experiments is depicted in Fig.~\ref{fig:waco}.
The hardware configuration consists of a variety of stock components, and custom built installations, thus forming a research platform suitable for the testing of software solutions for highly accurate mobile manipulation.
The robotic platform comprises a mobile base (Inspector Bots \emph{Super Mega Bot}\footnote{\href{www.inspectorbots.com}{www.inspectorbots.com}}) with mounted Kinova Jaco robot arm\footnote{\href{www.kinovarobotics.com}{www.kinovarobotics.com}}, a custom 3D-printed end-effector, sensor suite for localization, and computing hardware. The robot is named \emph{Waco}, derived from "Wheeled Jaco".
As a tool, the end-effector is equipped with a small spring in series with a marker, in order to perform non-destructive discrete tool positioning tasks.
For on-board localization, the robotic platform is equipped with a Velodyne VLP-16 sensor, an Xsens MTi-100 IMU, wheel encoders, and a sensor-head consisting of three orthogonal laser distance measurement sensors at the end-effector.
The on-board computing hardware is based on an Intel Core i7-6700 CPU @ \SI{3.4}{\giga\hertz}, and \SI{16}{\giga B} RAM. A second off-board computer, communicating over WLAN, runs the building task interface for the operator.
\setlength{\textfloatsep}{5pt}%
\subsection{Experimental setup}
\begin{figure}
    \centering
    \includegraphics[width = 0.9\columnwidth]{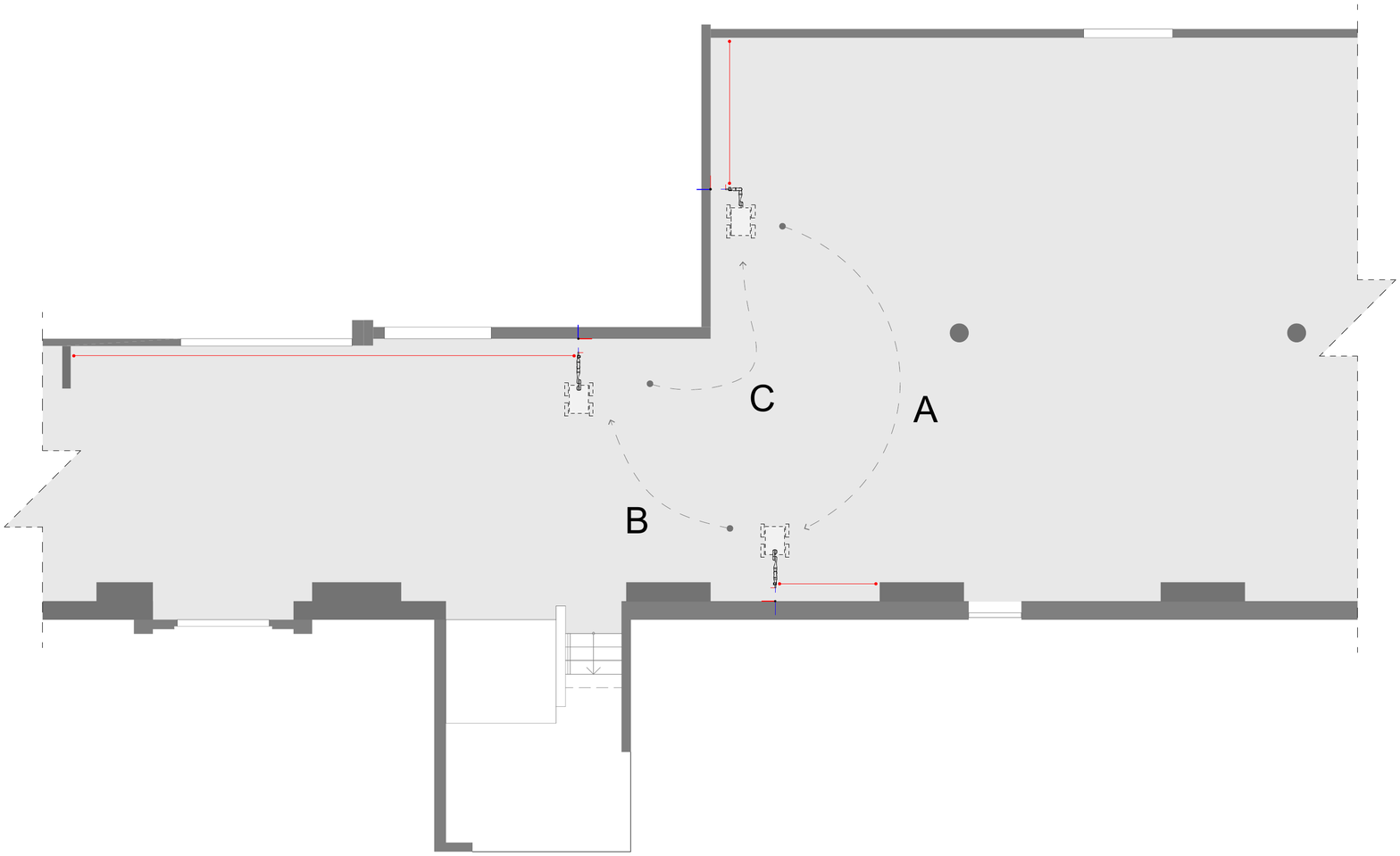}
    \vspace{-1.6em}
    \caption{Plan view of the experimental setup: The robot is illustrated in the three task locations A, B, and C. The red lines indicate the lateral \ac{HAL} references, the vertical reference is measured against the floor.}
    \vspace{-0.5em}
    \label{fig:plan}
\end{figure}

The system is tested in a realistic construction site environment with unfinished concrete walls, and some clutter, see Fig.~\ref{fig:teaser}, and Fig.~\ref{fig:plan}.
Using the building task interface, we specify interaction tasks on three different walls, which are the locations to be marked by the custom-made tool.
The system accuracy is first tested by repeatedly commanding different targets in a $3\times3$-dot pattern on each wall, as illustrated in Fig.~\ref{fig:pattern}. The robot then autonomously approaches the task locations from random locations several meters away, and places a single dot. This procedure is thereafter repeated. For later ground-truth evaluation, the commanded task locations and the final placements of the end-effector are measured with a Leica Nova TM50 Total Station.
Furthermore, we demonstrate the entire system performing fully autonomous loops between three task locations (one target on each wall), as depicted in Fig.~\ref{fig:teaser} and Fig.~\ref{fig:plan}.
Moreover, we vary the experimental setup by dynamically adding obstacles in the planned path of the robot, demonstrating adaptive re-planning of the platform.
In the experiments, we use a building model of the environment with minor deviations between the as-planned and as-built status.
However, notable deviations between the reality and the building plan persist, such as smooth surfaces instead of brick walls, unmodeled HVAC components such as pipes, and small deviations in walls.
\subsection{Results}
The integrated system experiment shows reliable performance throughout 30 individual experiments. The robot concludes multiple cycles between the three task locations without human intervention, also adaptively avoiding dynamic obstacles~\footnote{Please refer to the accompanying video for a full demonstration: \url{https://youtu.be/Ol82Gh_1T9w}.}.
\subsubsection{Positioning evaluation}
\begin{figure}
    \centering
    \includegraphics[width = 0.93\columnwidth]{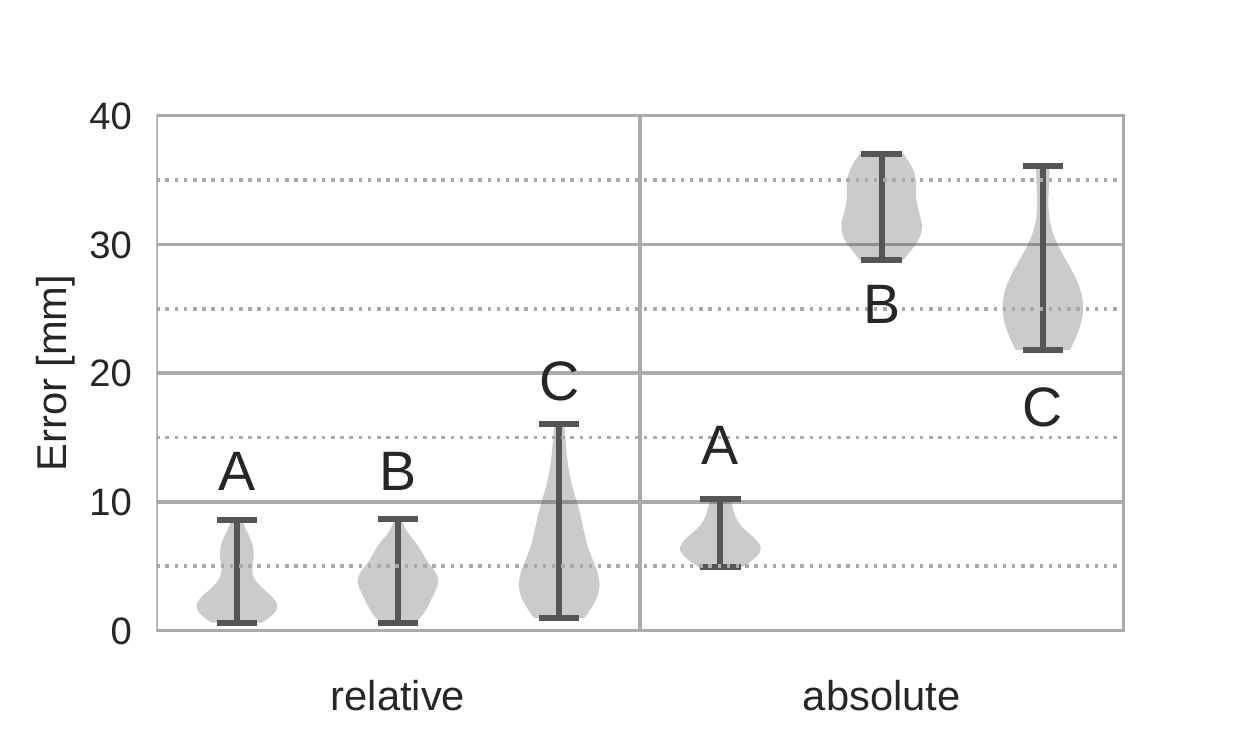}
    \vspace{-0.5em}
    \caption{Absolute and relative errors in point position for a nine dot pattern at the three different task locations.}
    \label{fig:all_errors}
    \vspace{-1.1em}
\end{figure}

\begin{figure}
    \centering
    \includegraphics[width = 0.93\columnwidth]{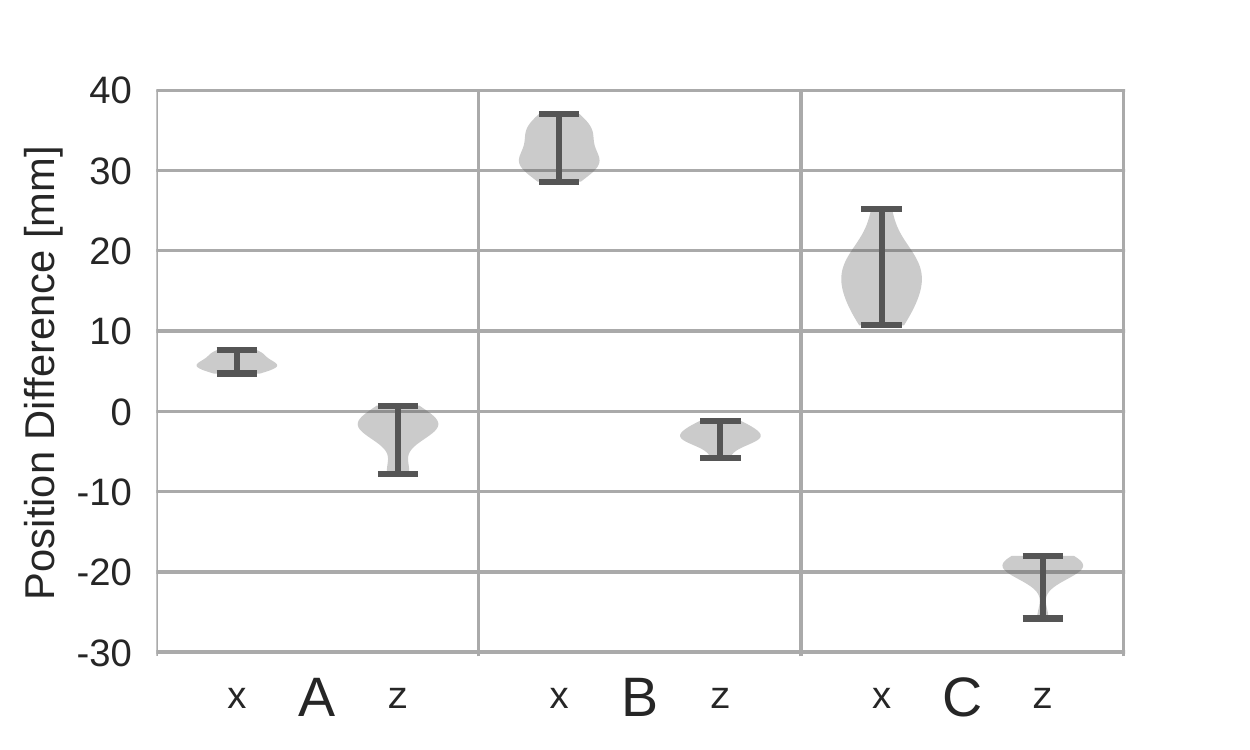}
    \vspace{-0.5em}
    \caption{Difference between commanded and executed point position over the different axes in the nine dot pattern. }
    \label{fig:abs_errors}
\end{figure}

\begin{figure}
    \centering
    \includegraphics[width = 0.8\columnwidth]{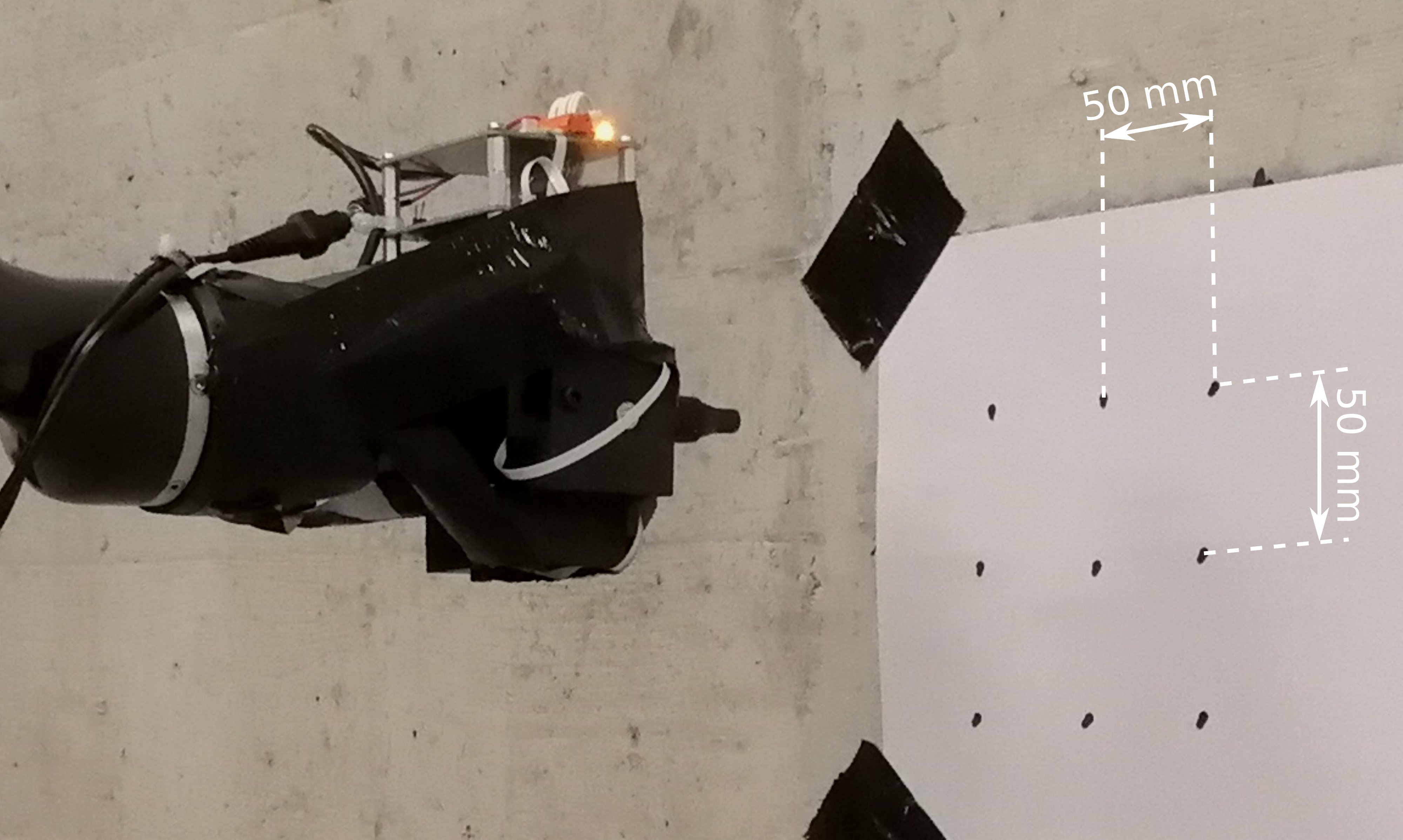}
    \vspace{-0.5em}
    \caption{Nine dot task pattern at location B: The robot is commanded to perform a \SI{50}{\milli\meter}-spaced $3\times3$-grid of interactions in three locations.}
    \label{fig:pattern}
    \vspace{-1.0em}
\end{figure}

The robot performance in the interaction experiment at the three task locations is reported in Fig.~\ref{fig:all_errors} and Table~\ref{tab:errors}. Here, we report the absolute errors in global coordinates, with respect to the building model origin used to derive the robot tasks. Furthermore, the relative errors within the drawing pattern of the robot are depicted. For this, all pair-wise combinations are evaluated. The plots show the distribution and total spread of all measurements. Overall, we observe mean relative errors between \SI{3.3}{\milli\meter} and \SI{5.9}{\milli\meter}. The absolute errors for the three task locations are below \SI{1}{\centi\meter} for location A, and between \SI{22}{\milli\meter} and \SI{37}{\milli\meter} for locations B and C respectively. The lateral (x) and vertical components (z) of the errors are split in Fig.~\ref{fig:abs_errors}. Notable observations are a high lateral offset for locations B and C, as well as a high vertical offset at task location C, while both error components are low at location A.
A sample $3\times3$ dot pattern created by the robot at location B is depicted in Fig.~\ref{fig:pattern}.

\setlength{\textfloatsep}{9pt}%
\begin{table}[]\caption{Mean positioning errors.}
\vspace{-0.5em}
    \centering
\begin{tabular}{lccc}
\toprule
{} &   A &    B &    C \\
\midrule
Mean absolute error [mm] & 7.0 & 32.9 & 26.1 \\
Mean relative error [mm] & 3.3 &  3.8 &  5.9 \\
\bottomrule
\end{tabular}
    \label{tab:errors}
\end{table}
\subsubsection{Discussion}
This section offers a discussion of the experimental results. Firstly, it is noteworthy that the system achieves generally low relative errors facilitating locally repeatable references. This is especially important for locally accurate construction tasks, e.g., installation of several screws in a specific pattern to mount a lamp or rail. However, the absolute error is presently highly dependent on the task location. While the principal solution of the \ac{HAL} routine is theoretically sound, we identified sources of error leading to increased absolute errors particularly for locations B and C.

In both locations, we note a large lateral absolute error. This can be best attributed to our current experimental set-up of taking reference walls with large distance in lateral direction as localization reference for the \ac{HAL} routine. Here, the distances between task locations, and reference walls are \SI{1.7}{\metre}, \SI{11.5}{\metre}, and \SI{4}{\metre} for locations A-C respectively. Slight rotational errors in the external calibration of the laser distance sensors are magnified by these distances. Furthermore, the floor at location C has an inclination that is not accurately reflected in the building model, adding error components to the absolute error both laterally, and vertically. Also, the laterally referenced walls for B and C are rough brick walls, instead of the load-bearing concrete walls in the building plan.

Lastly, we also tested the system with a simplified building model of the environment having several significant deviations from the as-built conditions, notably in the inclination of the floor, and the distances between the walls. Testing in this scenario led to a large increase in errors especially in relative errors for location C by factors of $2-3$, as most measurements were rejected as outliers given the discrepancy from the as-planned building model.
While the system performs well in locations of the building model that reflect the real conditions, a promising avenue of future research is handling of discrepancies between as-planned and as-built status of construction environments as well as handling of highly cluttered environments containing a large amount of unmodelled objects.

\section{Conclusions}
\label{sec:conclusions}
In this work, we presented a system for high-accuracy interactions of a mobile robot for construction sites. The system performs all state estimation using on-board sensing only. Notable features are a tight integration with the digital building model for task allocation, and robot localization. It is implemented on a robotic platform, and is able to reliably reach high-accuracy interactions under realistic conditions.
Presently, the \ac{HAL} routine is hand-designed. In future work it would be interesting to investigate task-adaptive strategies to yield the most reliable measurements. Furthermore, we wish to close the feedback loop between the perception and building task interface, automatically updating the building model perceived from the as-built conditions.


\bibliographystyle{IEEEtran}
\bibliography{references}

\begin{acronym}
\acro{DoF}{Degrees of Freedom}
\acro{HAL}{High Accuracy Localization}
\end{acronym}

\end{document}